# Multi-label ensemble based on variable pairwise constraint projection

Ping Li[*a, b], Hong Li[b], Min Wu[b]

[a]*College of Computer Science, Zhejiang University, Hangzhou 310027, China*
[b]*School of Information Science and Engineering, Central South University, Changsha 410083, China*

**Abstract:**
Multi-label classification has attracted an increasing amount of attention in recent years. To this end, many algorithms have been developed to classify multi-label data in an effective manner. However, they usually do not consider the pairwise relations indicated by sample labels, which actually play important roles in multi-label classification. Inspired by this, we naturally extend the traditional pairwise constraints to the multi-label scenario via a flexible thresholding scheme. Moreover, to improve the generalization ability of the classifier, we adopt a boosting-like strategy to construct a multi-label ensemble from a group of base classifiers. To achieve these goals, this paper presents a novel multi-label classification framework named *Variable Pairwise Constraint projection for Multi-label Ensemble* (VPCME). Specifically, we take advantage of the variable pairwise constraint projection to learn a lower-dimensional data representation, which preserves the correlations between samples and labels. Thereafter, the base classifiers are trained in the new data space. For the boosting-like strategy, we employ both the variable pairwise constraints and the bootstrap steps to diversify the base classifiers. Empirical studies have shown the superiority of the proposed method in comparison with other approaches.

## 1 Introduction

Traditional supervised learning deals with the data sample associated with only single label, which indicates its class membership. However, in a broad range of real-world applications, one sample is usually related to multiple class labels simultaneously, and such data has appeared in a large volume of various domains, e.g., scene or sentiment classification, video annotation and functional genomics [4, 10, 19, 25, 43]. To address this problem, multi-label learning has received a growing interest in the last decade [12, 30, 38]. Compared to single-label learning, multi-label learning is more challenging as each sample contains more than one label, e.g., a document concerning health might also be related to other topics, such as education, entertainment and government [8, 9, 21]. Typical multi-label learning methods include multi-label neural networks, multi-label transfer learning and multi-label kernel learning [13, 16, 34, 46], which are fundamentally derived from the corresponding single-label learning approaches. For more details, we refer the readers to [35].

In this paper, we study supervised multi-label classification. During the past few years, many multi-label classification methods have been proposed in a number of works [4, 9, 11, 27, 28], whereas they often neglect the pairwise relations suggested by sample labels. Such relations can indicate whether a given sample pair has similar label sets or not, which are of vital importance in multi-label classification. Besides, the pairwise label constraints have been proved to be effective in single-label

---

learning [51]. Therefore, we naturally extend the traditional pairwise constraints to the multi-label scenario via a flexible thresholding strategy. On the other hand, ensemble learning which combines multiple base learners to jointly accomplish one common task is shown to be very beneficial for enhancing the generalization ability of a single classifier [3, 11, 20, 24, 28, 32, 36, 44]. Moreover, ensemble learning is able to handle the imbalanced data well and discourage the over-fitting incurred by singular problems [22, 31, 33, 34]. Inspired by this, we adopt a boosting-like strategy to construct a multi-label ensemble, so as to further improve the generalization ability of the classifier. To achieve these goals, herein we propose a novel multi-label classification framework named *Variable Pairwise Constraint projection for Multi-label Ensemble* (VPCME) to deal with multi-label data. As a matter of fact, our framework can be mildly decomposed into two components, i.e., the variable pairwise constraint projection and the boosting-like process.

The main contributions of this work are highlighted as follows.

- We present a novel multi-label classification framework to construct a multi-label ensemble. It takes advantage of the variable pairwise constraint projection to obtain a well preserved low-dimensional data representation, where the base classifiers are learned.
- A boosting-like strategy is utilized to obtain a group of diversified base classifiers, whose diversities are essentially encouraged by both the variable pairwise constraints and the bootstrap steps. The attained base classifiers are combined to construct a multi-label ensemble for final classification.
- Extensive experiments were conducted on various real-life datasets. Results demonstrate that our framework enjoys significant advantages compared to other methods.

The remainder of the paper is structured as follows. We briefly review the multi-label classification and ensemble learning in Section 2. We introduce our multi-label classification framework as well as some rigorous analysis in Section 3. Experimental results on a broad range of real-world multi-label datasets are reported in Section 4. Finally, we provide the concluding remarks and discuss future work in Section 5.

## 2 Related works

Our work focuses on multi-label classification and the proposed framework is closely related to ensemble learning. So in this section, we provide a brief description about multi-label classification and ensemble learning.

### 2.1 Multi-label classification

Multi-label classification has received much attention in the past few years since it is practically relevant and interesting from a theoretical view [4, 10, 45, 48]. Traditional supervised classification concentrates on single-label data associated with only one label. When it is referred to binary classification, the cardinality of the discrete label set $|L|$ equals 2. When $|L|$ scales to more than two, it is treated as a multi-class problem. However, in many real-world applications, a large number of instances are usually correlated with multiple labels, which is called multi-label problem [34, 35]. Since problems of this type are ubiquitous, a great number of multi-label classification methods are continuously developed to deal with multi-label data from a broad range of areas.

In general, the existing multi-label classification techniques can be divided into two groups: *problem transformation* and *algorithm adaptation*. The former aims to transform multi-label problem into several single-label problems, e.g., Binary Relevance (BR) and Label Power-set (LP) [35]. The latter attempts to generalize existing single-label classification algorithms to multi-label cases, e.g., Boostexter [30]. BR learns *q* binary classifiers and one for each different label in *L*. LP regards each unique set of labels as one class for a new single-label classification task. *Zhang et al.* proposed the *Multi-Label K-Nearest Neighbor* (MLKNN) method [47], which implements the single classifier through the combination of KNN and Bayesian inference. Actually, it has been applied to many practical tasks due to its promising results as well as its simplicity.

Since multi-label data contains multiple labels, it is important to explore the interdependencies among labels and samples. For example, *Cheng et al.* come up with an approach based on a unified framework called IBLR-ML, which combines the model-based and similarity-based inference for multi-label classification [9]. In particular, IBLR regards the information from instances similar to the query as one additional feature, and treats the instance-based classification as the logistic regression. *Fürnkranz et al.* put forward a *Calibrated Label Ranking* (CLR) method for multi-label classification [11], which considers the calibrated scenario and extends the common learning using the pairwise comparison to the multi-label case. Besides, it makes use of an artificial calibration label to discriminate the relevant labels from the irrelevant ones for each sample. Recently, *Chen et al.* attempted to discover the multi-label temporal patterns in the sequence database and they treated events as multi-label ones with many states [8]. Furthermore, *Han et al.* presented a general sparse representation framework for multi-label transfer learning [13]. This approach learns a multi-label encoded sparse linear embedding space from a relevant dataset and then maps the target data into the new data space. In addition, to explore the tag account and the correlation information jointly, *Lo et al.* modeled the audio tagging task as a cost-sensitive multi-label learning problem [19].

## 2.2 Ensemble learning

Ensemble learning has established itself as a powerful learning paradigm in machine learning community, and received lots of interests in the last decade [3, 18, 23, 33, 39]. In this paradigm, a group of base learners are combined to construct an ensemble for one target problem. It is widely accepted that combining multiple classifiers should improve the generalization ability of the system under the assumption that the base learners are as accurate and diverse as possible [17]. Since identical base classifiers will not provide any additional information, it is desirable to diversify them in several alternative manners, e.g., sub-resampling training data, feature subsets selection, adding stochastic factors to the learning algorithm [23, 33]. Roughly speaking, the mainstream ensemble methods include *bagging* [2, 5], *boosting* (e.g., *AdaBoost*) [31, 37], *random forest* [6], *random subspace* [14] and *rotation forest* [29].

*Bagging* generates replicated training sets by sampling with replacement, and combines their classification results [2]. *Boosting* is more complex in the sense that the distribution of the training data is changing with the sequentially constructed classifiers, and it emphasizes more on the misclassified instances [31]. For *random subspace*, base classifiers are constructed in random subspaces [14]. In comparison, *bagging* can reduce the variation of the learner, which is more suitable for the decision tree and the neural networks; *boosting* is able to reduce the variation and bias at the same time, which is more appropriate for weak learners. In addition, *Rotation forest* is based on feature extraction, where features are randomly split into *K* subsets and then Principle Component Analysis (PCA) is used to reduce the dimension [29]. Thus, *K* axis rotations represent new features for base classifiers.

Besides the above classical methods, other ensemble methods such as *decision tree* ensemble [1], *active* ensemble [22], *dynamic* ensemble selection [43], *feature sets* ensemble [44] and *artificial neural network* ensemble [49], all targets on the single-label problem. Compare to them, multi-label ensemble is more complicated since it takes into account correlations among multiple labels. To this end, there are only a few works concerning this topic. For instance, *Zhang et al.* presented an ensemble method based on twin-SVM for multi-class and multi-label text categorization [50]. *Tsoumakas et al.* proposed a *RAndom K-labEL sets* (RAKEL) algorithm to construct individual ensemble learners [36], which draws a random subset of size *k* from all labels and the member of each ensemble constructs a *Label Power-set* classifier using a small random subset. Furthermore, *Read et al.* put forward a *Pruned Sets* method that only focuses on the most important label correlations [27] and proposed the *Classifier Chains* (CChains) to model label correlations with the tolerable computational complexity [28].

## 3 The VPCME method

In this section, we introduce our proposed VPCME framework. First, we give an explicit definition about the variable pairwise constraints. Second, the two inherent components of the multi-label classification framework are described, i.e., the variable pairwise constraint projection and the boosting-like strategy. Finally, we summarize our algorithm.

### 3.1 Variable pairwise constraints

Pairwise constraints are often used to characterize the relations among labels in many tasks, such as semi-supervised clustering [7, 52]. Traditional pairwise constraints are exclusively utilized for single-label problem and none works have applied this concept to multi-label problem to the best of our knowledge. In this work, we aim to explore the multi-label pairwise constraints to discover the latent relations among multiple labels. To achieve this goal, we propose a new concept called *variable pairwise constraints*, which is an extension of traditional ones. Details are shown as follows.

Recall that traditional pairwise constraints in single-label problem are generally defined as follows: If a sample pair shares the identical label, we impose the *must-link* constraint on them; otherwise impose the *cannot-link* constraint on them. Such constraints are called *fixed pairwise constraints*. However, when it comes to multi-label problem, fixed constraints become improper in the sense that too strict constraints will lead to serious unbalances between the *must-link* and *cannot-link* constraints. Worse still, once constraints are fixed, it might be impossible to further diversify the base classifiers. As a matter of fact, these shortcomings would bring about many inconveniences with regard to the succeeding tasks, e.g., constraint projection and multi-label classification. Therefore, it is very natural for us to extend the traditional pairwise constraint to multi-label scenarios, i.e., the *variable pairwise constraints*. Since multi-label data are related to multiple labels, we give an intuitive definition that if the percentage of the same labels in a sample pair is larger than or equal to some threshold, then this sample pair will be imposed the *must-link* constraint; otherwise the *cannot-link* constraint. In reality, sample pairs with *must-link* constraints form a *must-link* set and sample pairs with *cannot-link* constraints form a *cannot-link* set.

Hopefully, the performance of multi-label classification algorithms could be improved by taking into account the inherent correlations among multiple labels. In our framework, the label correlations are reflected in terms of variable pairwise constraints. Concretely, the *must-link* set collects the sample pairs with more similar labels during re-sampling process. Meanwhile, the *cannot-link* set collects sample pairs with more dissimilar labels. These two kinds of label relationships will be well preserved via the variable pairwise constraint projection as shown in the following part.

## 3.2 Variable pairwise constraint projection

It is clear that one promising multi-label ensemble requires that the base classifiers should be as accurate and diverse as possible. This part focuses on obtaining a well preserved lower-dimensional data representation using the variable pairwise constraint projection. The base classifier is learned in the new data space to capture better discriminating power. We begin with the mathematical definitions described as follows.

Assume that a given multi-label dataset consists of $n$ samples with $r$ classes. Each sample has $k$ features stacked in a row vector. The feature matrix is denoted by $X = [x_1, x_2, \ldots, x_n]^T$, and the label matrix is $Y = [y_1, y_2, \ldots, y_n]^T$, where $y_i = [y_{i1}, y_{i2}, \ldots, y_{ir}]$ (using PT3 [32]). If the sample $x_i$ in X can be categorized to the $r$th class, then $y_{ir} = 1$, otherwise $y_{ir} = -1$. Our goal is to precisely estimate the label sets of test data by using the proposed multi-label classification framework.

As mentioned earlier, it makes sense that we learn a lower-dimensional data representation through the variable pairwise constraint projection, which can preserve the inherent correlations among multiple labels. Now, we formulate the two kinds of variable pairwise constraints according to the definition in Section 3.1 as follows:

$$M = \{(x_i, x_j) \mid |R(x_i) \cap R(x_j)| / [(|R(x_i)| + |R(x_j)|)/2] \geq \Theta\} \tag{1}$$

$$C = \{(x_i, x_j) \mid |R(x_i) \cap R(x_j)| / [(|R(x_i)| + |R(x_j)|)/2] < \Theta\} \tag{2}$$

where M denotes the *must-link* set, C denotes the *cannot-link* set, $\Theta$ is a constant ranging from 0 to 1, $R(\,.\,)$ is the label set of a sample, and $|R(\,.\,)|$ is the cardinality of the label set (i.e., the number of elements in the set). Besides, $|R(x_i) \cap R(x_j)|$ represents the number of the same labels shared by the data pair $(x_i, x_j)$. In principle, the threshold $\Theta$ is used to tradeoff the balance between the *must-link* set and the *cannot-link* set, which has large influences on the new data representation.

In order to obtain the low-dimensional data representation, we follow [51] and seek a group of projection vectors $W = [w_1, w_2, \ldots, w_d]$ that best preserve the correlated pairwise constraints of M and C in the new data space. Thus, we can derive the low-dimensional data representation from the transformation $z_i = W^T x_i$. Ideally, we expect that the two samples of any data pair in M are as close as possible and the two samples of any data pair in C are as far apart as possible. Therefore, we want to maximize the objective function formulated as:

$$J(W) = \frac{1}{2n_C} \sum_{(x_i,x_j) \in C} \|W^T x_i - W^T x_j\|^2 - \frac{r}{2n_M} \sum_{(x_i,x_j) \in M} \|W^T x_i - W^T x_j\|^2, \ s.t. \ W^T W = I \tag{3}$$

where $n_C$ and $n_M$ denote the sizes of C and M respectively. Practically, $n_C$ and $n_M$ can be varied when necessary. In this work, they are set to the number of instances. The term $r$ is a scaling coefficient, which is utilized to govern the contribution of data pairs in M to the objective function. Since the distance between samples in M is typically smaller than that in C, then $r$ can be estimated by:

$$r = \left[\frac{1}{n_C} \sum_{(x_i,x_j) \in C} \|x_i - x_j\|^2\right] \bullet \left[\frac{1}{n_M} \sum_{(x_i,x_j) \in M} \|x_i - x_j\|^2\right]^{-1} \tag{4}$$

After some linear algebraic operations, the objective $J(W)$ in Eq.(3) can be simplified as the matrix trace, i.e., $Tr(W^T (S_C - rS_M) W)$. We assume that a group of vectors $\{W_1, W_2, \ldots, W_d\}$ are the eigenvectors corresponding to the $d$ largest eigenvalues $(\lambda_1, \lambda_2, \ldots, \lambda_d)$ of the matrix $S_C - rS_M$, and $\wedge$ is a diagonal matrix with the eigenvalues as its diagonal elements. Then $Tr(W^T (S_C - r S_M) W)$ achieves the optimal value when the number of eigenvectors $d$ is set to the number of non-negative eigenvalues [51], which is also the dimension in the new data space. Thus, $J(W)$ can be rewritten as:

$$J(W) = Tr(W^T(S_C - rS_M)W) = Tr(\Lambda) = \sum_{i}^{d} \lambda_i \quad (5)$$

where $S_C$ is the *cannot-link* scatter matrix and $S_M$ is the *must-link* scatter matrix, shown by:

$$S_C = \frac{1}{2n_C} \sum_{(x_i, x_j) \in C} (x_i - x_j)(x_i - x_j)^T \quad (6) \qquad S_M = \frac{1}{2n_M} \sum_{(x_i, x_j) \in M} (x_i - x_j)(x_i - x_j)^T \quad (7)$$

Now, we can obtain the new data representation $(z_i, y_i)$, where the base classifier is learned. Note that the above formulations are similar to those in [51], which is only designed for single-label problem. In contrast, our framework aims to solve multi-label problem and takes into account the correlations among multiple labels.

### 3.3 The boosting-like strategy

In section 3.2, we can obtain the new data representation to learn an accurate classifier, but it suffers from the weak generalization ability. To overcome this drawback, we adopt a boosting-like strategy to obtain a group of diverse base classifiers in this section.

Different from *bagging* or *boosting* which resample the training data directly [2, 5], we resample pairwise constraints, which simultaneously considers the features and the label sets. By using the boosting-like strategy, the selected pairwise constraints are changing with a flexible thresholding scheme, thus providing diverse information as much as possible for the variable pairwise constraint projection. This way, the base classifiers learned in the new lower-dimensional data space are expected to become as diverse as possible.

On the other hand, in our framework we also use a technique similar to *boosting* [31], i.e., emphasizing more on misclassified samples. Here, the iteration is defined as one bootstrap step, which provides further diversity for the base classifier. Particularly, assume that each sample is initially assigned the same weight $w_i = 1/n$, we will endow the misclassified samples with an increased weight $w_i(1 + \theta)$ in iteration, where $\theta$ is a weight scaling factor, e.g., it can be the training error rate in previous iteration. Note that the correctly classified samples remain their weights during the process.

In this boosting-like strategy, we cooperatively take advantage of the variable pairwise constraints and the bootstrap steps to diversify the base classifiers, such that we can obtain a collection of base classifiers as diverse as possible.

### 3.4 The VPCME framework

In this part, we summarize the proposed multi-label classification framework named *Variable Pairwise Constraint projection for Multi-label Ensemble*. The pseudo-codes of the sketch are shown in Table 1. In the following, we elaborate this framework more clearly.

First, we randomly select the data pairs $(x_i, x_j)$ from the training data set and put them into the corresponding constrained set $C$ or $M$ according to the flexible thresholding scheme as described in Section 3.1. In total, we select $n_C$ samples from the *cannot-link* set C and $n_M$ samples from the *must-link* set M. Second, we obtain a new data representation with the reduced dimension $d$ by using the variable pairwise constraint projection, i.e., $Z = W^T X$. Third, the base multi-label classifier is learned in the new data space. Furthermore, we adopt the boosting-like strategy to repeat the above process until the desired number of base classifiers is achieved. Ultimately, a collection of base classifiers are combined together to construct a robust multi-label ensemble. For prediction, we simply use the popular *majority voting* to estimate the label sets of the test data, since this method does not require any prior field knowledge and has low computational cost [23, 24].

Table 1 The pseudo-code of the proposed VPCME framework

| | Algorithm: VPCME |
|---|---|
| 0 | **INPUT:** |
| 1 | Training data $(x_i, y_i)$, $x_i = [x_{i1}, x_{i2},…, x_{ik}]^T$, $y_i = [y_{i1}, y_{i2},…, y_{ir}]$, $k$ is the number of attributes, $r$ is the |
| 2 | cardinality of label sets in total, base classifier $BC$, ensemble size $S$, variable pairwise constraints sets |
| 3 | $C$ and $M$ with size $n$, the threshold $\Theta$, initial weights $w_i$. |
| 4 | **INITIALIZE:** $ind(x)=\{1,2,…,n\}$, $C = \Phi$(empty set), $M = \Phi$, $w_i = 1/n$. |
| 5 | **FOR** $l = 1, 2, …, S$ (the number of base classifiers) |
| 6 |   **WHILE** $|C| <= n_C$ and $|M| <= n_M$ |
| 7 |     Randomly select one pair of samples indexed by $i$ and $j$ from $(x_i, y_j)$, where $i$ does not equal $j$. |
| 8 |     **IF** $|R(x_i) \cap R(x_j)| / [(|R(x_i)| + |R(x_i)|) / 2] >= \Theta$ |
| 9 |       Add the data pair $(x_i, x_j)$ to the *must-link* set $M$. |
| 10 |     **ELSE** |
| 11 |       Add the data pair $(x_i, x_j)$ to the *cannot-link* set $C$. |
| 12 |     **ENDIF** |
| 13 |   **ENDWHILE** |
| 14 |   **a.** Compute the projection matrix $W_l$ using the *must-link* set $M$ and the *cannot-link* set $C$. |
| 15 |   **b.** The new data representation $(z_i, y_i)$ can be obtained by the matrix transformation $z_i = W^T x_i$. |
| 16 |   **C.** Learn the individual classifier $BC_l$, and update corresponding weights like $w_i \leftarrow w_i(1+\theta)$ for the |
| 17 | misclassified samples $(z_i, y_i)$, the term $\theta$ equals the misclassification rate in the previous iteration. |
| 18 | **ENDFOR** |
| 19 | **OUTPUT:** A group of generated base classifiers ($|S|$) jointly predicting the label set of the testing |
| 20 | samples using majority voting. |

Essentially, our framework consists of two components, i.e., the variable pairwise constraint projection and the boosting-like strategy. Both of them are cooperatively applied to construct a robust multi-label ensemble, e.g., both the variable pairwise constraints and the bootstrap steps are used to diversify the base classifiers. Notice that the constrained projection is closely related to dimensionality reduction [15, 26, 38, 53], but they are applied differently. For one thing, our projection is based on the variable pairwise constraint sets (i.e., $M$ and $C$) rather than the original data set. For another, the boosting-like strategy together with the variable pairwise constraint projection jointly contributes to obtain multiple base classifiers as accurate and diverse as possible, which is not the case for those feature extraction methods. In addition, this projection could preserve the pairwise relations between multiple labels and samples, which are beneficial for training desirable base classifiers.

## 4 Experiments

This section show empirical studies on a broad range of real-world multi-label datasets. First, we give a brief description about the evaluation metrics and data sets. Second, experimental setup and performance comparisons are presented. Third, we report the results as well as some analysis.

### 4.1 Evaluation metrics

In this test, we employ several popular metrics to evaluate the proposed framework [35, 47, 48]. Suppose that a multi-label dataset $D$ consists of $N$ instances, represented by $D = (x_i, Y_i)$, $i =1, … , N$, where $Y_i \subseteq L$ is the true label set and $L=\{\lambda_j : j =1, \ldots , M\}$ is the total label set. For a given instance $x_i$, its estimated label set is denoted by $Z_i$, and the estimated rank of the label $\lambda$ is denoted by $r_i(\lambda)$. The most relevant label takes the top rank (1) and the least one only gets the lowest rank ($M$). In the following, we explain these evaluation metrics from a mathematical viewpoint.

*Hamming loss* was initially proposed by *Schapire* and *Singer* [30], and it enumerates the misclassified times of the predicted labels based on instances. It can be defined as follows:

$$HammingLoss = \frac{1}{N}\sum_{i=1}^{N}\frac{|Y_i \Delta Z_i|}{M} \quad (8)$$

where the symbol "$\Delta$" stands for the symmetric difference of the two sets $Y_i$ and $Z_i$, and that is the set-theoretic equivalent of the exclusive disjunction (XOR operation) in Boolean logic.

*Ranking loss* denotes the number of times that irrelevant labels are ranked higher than relevant labels, and it takes the form:

$$RankingLoss = \frac{1}{N}\sum_{i=1}^{N}\frac{1}{|Y_i||\overline{Y_i}|}\left|\{(\lambda_a, \lambda_b) : r_i(\lambda_a) > r_i(\lambda_a), (\lambda_a, \lambda_b) \in Y_i \times \overline{Y_i}\}\right| \quad (9)$$

where the term $\overline{Y_i}$ is the complementary set of $Y_i$ with respect to $L$.

*One-error* mainly evaluates how many times the top-ranked label is not in the set of relevant labels of the instance. It can be formulated as follows:

$$OneError = \frac{1}{N}\sum_{i=1}^{N}\delta\left(\arg\min_{\lambda \in L} r_i(\lambda)\right), \text{ where } \delta(\lambda) = \begin{cases} 1, \text{ if } \lambda \notin Y_i \\ 0, \text{ otherwise} \end{cases} \quad (10)$$

*Coverage* examines how far we need on average to go down the rank list of labels in order to cover all the relevant labels of the instance. This can be expressed by:

$$Coverage = \frac{1}{N}\sum_{i=1}^{N}\max_{\lambda \in Y_i} r_i(\lambda) - 1 \quad (11)$$

*Average precision* assesses the average fraction of labels ranked above a particular label $\lambda \in Y_i$ which are actually in $Y_i$. And this metric virtually reflects the average classification accuracy of the predicted labels of the instance. It can be denoted by:

$$AvePrecision = \frac{1}{N}\sum_{i=1}^{N}\frac{1}{|Y_i|}\sum_{\lambda \in Y_i}\frac{\left|\{\lambda' \in Y_i : r_i(\lambda') \leq r_i(\lambda)\}\right|}{r_i(\lambda)} \quad (12)$$

Both of the two metrics *F1-metric* and *Recall* are elaborated by *Godbole & Sarawagi* [10], and they can be shown respectively as:

$$F1-metric = \frac{1}{N}\sum_{i=1}^{N}\frac{2|Y_i \cap Z_i|}{|Y_i|+|Z_i|} \quad (13) \qquad Recall = \frac{1}{N}\sum_{i=1}^{N}\frac{|Y_i \cap Z_i|}{|Y_i|} \quad (14)$$

In the above criteria, *Hamming loss*, *Ranking loss*, *One-error* and *Coverage* suffice to "*the smaller the better*" while *Average precision*, *F1-metric* and *Recall* characterizes "*the larger the better*". These metrics are employed jointly to investigate performances of multi-label classification methods [34, 53].

## 4.2 Datasets

To examine the multi-label algorithms, we compile a variety of multi-label datasets[1], including text categorization (*Yahoo data*, *enron*, *medical*) [47], image classification (*scene*) and bioinformatics (*yeast*, *genbase*) [35]. In summary, twelve datasets were used with 6 to 45 labels and from less than 700 examples to over 2, 400 ones. Their statistics are listed in Table 2.

---

[1] The Yahoo! datasets are available for download at http://www.kecl.ntt.co.jp/as/members/ueda/yahoo.tar.gz, and the remaining datasets can be obtained from http://mlkd.csd.auth.gr/multilabel.html

Table 2 Multi-label datasets and their statistics

| Dataset | domain | instances | features | labels | distinct | cardinality | density |
|---|---|---|---|---|---|---|---|
| yeast | biology | 2417 | 103 | 14 | 198 | 4.237 | 0.303 |
| scene | multimedia | 2407 | 294 | 6 | 15 | 1.074 | 0.179 |
| enron | text | 1702 | 1001 | 53 | 753 | 3.378 | 0.064 |
| genbase | biology | 662 | 1186 | 27 | 32 | 1.252 | 0.046 |
| medical | text | 978 | 1449 | 45 | 94 | 1.245 | 0.028 |
| arts | Yahoo text | 2000 | 462 | 26 | 254 | 1.627 | 0.063 |
| business | Yahoo text | 2000 | 438 | 30 | 96 | 1.590 | 0.053 |
| education | Yahoo text | 2000 | 550 | 33 | 200 | 1.465 | 0.044 |
| entertainment | Yahoo text | 2000 | 640 | 21 | 148 | 1.426 | 0.068 |
| health | Yahoo text | 2000 | 612 | 32 | 164 | 1.667 | 0.052 |
| science | Yahoo text | 2000 | 743 | 40 | 261 | 1.489 | 0.037 |
| social | Yahoo text | 2000 | 1047 | 39 | 137 | 1.274 | 0.033 |

The label *cardinality* is the average number of labels assigned to each sample, which is used to quantify the number of distinct labels. The larger the *cardinality*, the more difficult we obtain the impressive classification performance. Besides, this inherent property has direct impacts on *Coverage*. Specifically, two datasets with the identical *cardinality* but distinct label numbers might not exhibit the same property Label *density* is the percentage of the averaged labels in the total labels. The value of *density* associates with *Hamming loss* and *Ranking loss*, and it will affect the values of *F1-metric* and *Recall* as well. Label *distinct* denotes the number of different label combinations, and it is of key importance for many *algorithm adaptation* methods that operate on label subsets. The larger the *distinct* takes, the more complex the multi-label problem becomes. If some minority label combinations appear in an extremely small size, i.e., the imbalanced problem, *Coverage* will reduce.

The *yeast* and *genbase* datasets are both from the biological field. The *yeast* data is associated with 14 functional classes from the Comprehensive Yeast Genome Database of the Munich Information Center for Protein Sequences. The total number of genes amounts to 2417 with each gene represented by a 103-dimensional feature vector. In the *genbase* dataset, there are 27 primary protein families, including PDOC00064 and PDOC00154. After preprocessing, the data set consists of 662 proteins and each protein might belong to one or more of 27 classes.

In the multimedia domain, the *scene* dataset contains six possible scenes, such as *beach, sunset, field, fall foliage, mountain* and *urban*. It targets recognizing which of the above scenes can be observed in 2407 pictures. For a *scene* image, the spatial color moments are regarded as features and each picture is decomposed into 49 blocks using one 7 by 7 grid.

The *enron* dataset is collected and prepared by the Cognitive Assistant that Learns and Organizes (CALO) Project, containing data from about 150 users, mostly senior management of Enron, organized into folders. A subset of 1702 labeled email messages with 1001 characteristics is used in experiments and each message is labeled by two persons.

The *medical* data is compiled for the Computational Medicine Centers to challenge the international Natural Language Processing (NLP) community. In 978 documents, each involves a brief clinical free-text summary of the patient symptom history and their prognosis, labeled with insurance codes. They are associated with one or more labels from a subset of 45 candidate labels.

Seven *Yahoo!* datasets (i.e., *arts*, *business*, *education*, *entertainment*, *health*, *science* and *social*) are collected from the real Web pages linked to the "Yahoo.com" domain, including fourteen top-level categories and each category is divided into several second-level subcategories [47]. Focused on the second-level categories, 7 out of the 14 independent text categorization problems are considered here with each containing 2000 documents.

### 4.3 Experimental setup and design

In this part, we give descriptions about parameter settings and performance comparisons for our proposed framework.

### 4.3.1 Parameter settings

In the experiments, we compared VPCME with several state-of-the-art algorithms, including IBLR-ML [9], CLR [11], CChains [28], RAKEL [36] and MLKNN [47]. To investigate the individual components of VPCME, three typical ensemble methods are also examined, i.e., AdaBoost with pruned decision tree (AdaBoost$_{PDT}$), Bagging with the variable pairwise constraint projection (Bagging$_{VPCP}$), and multi-label dimensionality reduction via dependence maximization (MDDM) [53]. All experiments were carried out on a PC machine with Intel (R) Core (TM) Duo CPU 3.16 GHz and 2 GB RAM, Matlab R2009b version. The instance-based learning method MLKNN was used as the base classifier for VPCME due to its excellent predictive performance, and the number of nearest neighbor $k$ was set to 10 as in [47], where it was found to yield the most satisfactory performances.

To eliminate the bias incurred by different base learners, we respectively utilized KNN, C4.5 (i.e., J48 in WEKA [42]) and SMO for the compared algorithms RAKEL, IBLR-ML, CLR and CChains. Results show that KNN slightly degrades the performance and results of J48 and SMO are similar, so we only report the results of J48. For RAKEL, $k$ was set to |labels|/2 [36], and the smaller $k$ leads to the lower computational cost. Parameters for IBLR-ML, CLR, CChains were all established as described in original literatures [9, 11, 28]. The remaining parameters were set to their default values in *mulan*[2] [34] and WEKA [42] with Java JDK 1.6. Note that AdaBoost$_{PDT}$ handles one of the assigned multiple labels sequentially. The settings for Bagging$_{VPCP}$ were the same as those for VPCME except the bootstrap steps. The threshold for MDDM was set to 99% as in [53].

Since the generalization ability is of vital importance for one learning framework, we investigate the performance of our framework under various parameter settings in light of some refinement methods [40, 41], such as the threshold selection and different ensemble sizes. For all the tests, *five-fold* cross validations were carried out to estimate the labels. In detail, the original dataset is randomly divided into five parts with each almost the same size, and in each fold one of them is held out for testing and the remainder for training. This process was repeated five times so that each part would be treated as the test data exactly once. Without loss of generality, we repeated each test run for 20 times and recorded the averaged results as well as the standard deviations. Furthermore, the paired *t*-tests at the significance level of 0.01 were done to validate the efficacy of our approach.

---

[2] *Mulan* is available at http://mulan.sourceforge.net/download.html

Table 3(a) Performance of different algorithms in terms of *Hamming loss*. (mean±std. %)

| Dataset | RAKEL | IBLR-ML | CLR | CChains | MLKNN | VPCME |
|---|---|---|---|---|---|---|
| *yeast* | 23.15±0.76• | 19.37±1.04• | 22.02±0.84• | 26.71±0.59• | 19.29±1.05• | 17.57±0.18 |
| *scene* | 10.63±0.49• | 8.56±0.30• | 13.36±0.49• | 14.43±0.61• | 8.90±0.33• | 7.13±0.24 |
| *enron* | 4.81±0.19 | 5.61±0.17• | 4.70±0.14 | 5.27±0.14• | 5.30±0.15• | 4.42±0.15 |
| *genbase* | 0.15±0.03○ | 0.26±0.12 | 0.21±0.07○ | 0.15±0.03○ | 0.48±0.15 | 0.47±0.03 |
| *medical* | 1.01±0.11 | 2.07±0.03• | 1.04±0.11 | 1.02±0.11 | 1.55±0.18 | 1.25±0.08 |
| *arts* | 6.24±0.17 | 6.08±0.24 | 5.94±0.11 | 7.70±0.15• | 6.13±0.21 | 5.89±0.13 |
| *business* | 2.97±0.08 | 2.69±0.12 | 3.05±0.11• | 3.24±0.12• | 2.62±0.09 | 2.71±0.07 |
| *education* | 4.40±0.14• | 4.24±0.09• | 4.44±0.15• | 5.44±0.11• | 4.13±0.10 | 3.79±0.12 |
| *entertain.* | 6.11±0.16• | 6.01±0.25 | 6.13±0.16• | 7.05±0.35• | 5.94±0.26 | 5.62±0.14 |
| *health* | 3.83±0.12• | 4.29±0.22• | 4.03±0.08• | 4.28±0.21• | 4.43±0.23• | 3.36±0.11 |
| *science* | 3.89±0.11• | 3.83±0.04• | 3.65±0.05 | 4.40±0.19• | 3.63±0.03 | 3.42±0.06 |
| *social* | 2.40±0.09• | 2.89±0.14• | 2.44±0.09• | 2.62±0.09• | 2.69±0.09• | 2.10±0.04 |

### 4.3.2 Experimental design

In this subsection, we briefly describe four experimental groups of performance comparisons, which are designed to explore the performance of our VPCME method. Note that the remaining parameters of Group2, Group3 and Group4 are fixed as the same in Group1. Detailed descriptions are shown as follows.

**Group1:** To explore the performances of all compared multi-label classification algorithms, experiments on twelve datasets were conducted. For VPCME, the variable pairwise constraint threshold was empirically set to 0.6 and the ensemble size was tuned to 30. Evaluation results were recorded in terms of five metrics.

**Group2:** To exhibit the influences of different variable pairwise constraint thresholds on our framework, we tested an ascending threshold list ranging from 0.1 to 1.0 at the interval of 0.1 on *yeast* and *business*. This empirically provides an intuitive selection way for this parameter.

**Group3:** To investigate the performance of VPCME under different ensemble sizes, we examined a sequence of sizes from 10 to 50 with a grid of 10 on *medical* and *entertainment*. The larger the ensemble size, the more time complexity the multi-label classification method will cost. To attain satisfactory performance, a proper ensemble size is desired.

**Group4:** To examine the individual components of VPCME, we compared three other approaches, i.e., AdaBoost$_{PDT}$, Bagging$_{VPCP}$, MDDM. To ensure fairness, we tested on several representative datasets since they come from distinct domains, i.e., *yeast*, *scene* and *entertainment*.

### 4.4 Results

In this section, we report the results of four experimental groups in Section 4.3.2 respectively as well as some analysis. Table 3 (a-e) and Table 4 report the results of Group1, Fig. 1 depicts the results of Group2, the results of Group3 are tabulated in Table 5 (a-b), and the comparison results of Group4 are shown in Fig. 2.

Table 3(b) Performance of different algorithms in terms of *Ranking loss.* (mean±std. %)

| Dataset | RAKEL | IBLR-ML | CLR | CChains | MLKNN | VPCME |
|---|---|---|---|---|---|---|
| *yeast* | 22.36±1.10• | 16.50±0.99• | 17.81±1.03• | 32.61±0.79• | 16.60±0.95• | 12.91±0.52 |
| *scene* | 10.94±0.93• | 7.54±0.44• | 10.00±0.64• | 25.05±2.67• | 7.90±0.49• | 4.74±0.37 |
| *enron* | 20.35±1.11• | 10.69±0.81• | 7.28±0.77• | 18.44±0.56• | 9.37±0.68• | 4.59±0.67 |
| *genbase* | 0.29±0.29 | 0.56±0.28• | 1.25±0.69• | 0.35±0.25 | 0.81±0.38• | 0.20±0.05 |
| *medical* | 7.53±0.13• | 6.87±0.75• | 2.71±0.48• | 7.62±1.54• | 4.17±0.52• | 1.70±0.21 |
| *arts* | 27.90±0.90• | 17.15±0.60• | 12.91±0.90• | 24.63±1.11• | 15.89±0.27• | 9.14±0.36 |
| *business* | 11.68±0.98• | 4.64±0.61• | 3.64±0.42• | 11.19±1.04• | 3.95±0.38• | 1.78±0.25 |
| *education* | 34.94±1.81• | 11.07±0.58• | 8.63±0.65• | 23.85±1.59• | 8.97±0.72• | 5.45±0.27 |
| *entertain.* | 32.02±1.61• | 13.34±0.45• | 10.90±0.33• | 24.54±3.08• | 12.43±0.46• | 7.02±0.39 |
| *health* | 20.04±1.85• | 6.69±0.40• | 4.62±0.39• | 14.16±2.33• | 6.16±0.34• | 3.12±0.21 |
| *science* | 31.79±0.81• | 17.41±0.52• | 11.42±0.32• | 26.36±1.71• | 13.97±0.31• | 6.60±0.45 |
| *social* | 20.24±1.47• | 8.97±0.93• | 5.81±0.49• | 13.73±0.85• | 6.91±0.84• | 2.58±0.23 |

Table 3(c) Performance of different algorithms in terms of *One-error.* (mean±std. %)

| Dataset | RAKEL | IBLR-ML | CLR | CChains | MLKNN | VPCME |
|---|---|---|---|---|---|---|
| *yeast* | 29.33±2.39• | 22.13±2.14• | 23.83±1.66• | 34.26±1.10• | 22.88±1.91• | 18.56±0.76 |
| *scene* | 28.17±1.18• | 22.52±1.43• | 29.33±1.77• | 39.59±2.34• | 23.10±1.30• | 17.39±0.82 |
| *enron* | 29.14±4.81• | 37.02±3.10• | 21.74±2.26• | 42.36±1.21• | 31.20±2.54• | 20.35±0.79 |
| *genbase* | 0.76±0.68 | 1.66±1.47 | 0.15±0.30 | 0.45±0.37 | 0.76±1.17 | 1.09±0.67 |
| *medical* | 18.21±3.40• | 34.05±2.16• | 16.47±2.40 | 18.31±3.49• | 25.46±4.29• | 16.16±1.56 |
| *arts* | 58.75±1.44• | 62.60±2.57• | 52.95±1.44• | 63.90±2.08• | 62.35±2.11• | 42.74±0.82 |
| *business* | 14.85±1.01• | 12.00±1.64 | 12.60±1.81• | 23.00±0.85• | 11.65±1.25 | 11.24±1.03 |
| *education* | 59.80±1.23• | 60.45±0.87• | 57.35±0.94• | 66.30±0.81• | 58.20±1.44• | 45.30±0.35 |
| *entertain.* | 51.00±0.89• | 54.20±0.98• | 48.25±1.18• | 56.55±2.58• | 53.65±1.45• | 39.36±0.82 |
| *health* | 32.85±1.75• | 39.50±2.04• | 31.60±2.15• | 40.30±2.38• | 40.00±2.07• | 27.10±1.14 |
| *science* | 62.90±1.60• | 69.35±1.59• | 59.55±1.96• | 68.30±1.14• | 67.00±1.90• | 49.52±1.25 |
| *social* | 34.85±2.15• | 43.95±0.91• | 35.25±1.30• | 40.20±1.85• | 41.65±1.71• | 31.68±0.94 |

### 4.4.1 Results of Group1

Table 3 shows the results of different multi-label algorithms on several data sets. Records are tabulated in terms of averaged mean values as well as standard deviations over 20 test runs. To examine whether the results are statistically significant, paired t-tests were carried out at 1% significance level. The marker "•/○" suggests our approach is statistically superior/inferior to others. Note that the symbol "↓" indicates the smaller the better while "↑" indicates the larger the better. Specifically, when the presented method achieves significantly better/worse performance than the others, a win/loss is counted and a marker "•/○" is aside the record. Otherwise, a tie is counted and no marker is placed. The obtained win/tie/loss counts for VPCME against the compared algorithms are summarized in Table 4.

From Table 3 and Table 4, a number of interesting points can be observed as follows:

(1) Our VPCME approach systematically and consistently performs better than other algorithms, since it takes advantage of variable pairwise constraint projection and the boosting-like strategy. Particularly, the misclassified samples will receive more emphasis in iteration. Multiple diversified base classifiers are combined to construct a robust multi-label ensemble, which is able to achieve a small error rate by utilizing the additional information provided by different base classifiers.

Table 3(d) Performance of different algorithms in terms of *Coverage*. (mean±std. %)

| Dataset | RAKEL | IBLR-ML | CLR | CChains | MLKNN | VPCME |
|---|---|---|---|---|---|---|
| *yeast* | 7.67±0.20• | 6.22±0.12 | 6.72±0.16• | 8.94±0.09• | 6.24±0.13 | 5.94±0.08 |
| *scene* | 0.64±0.05• | 0.46±0.01 | 0.58±0.03• | 1.36±0.12• | 0.48±0.02 | 0.30±0.05 |
| *enron* | 25.39±0.59• | 14.89±0.73• | 11.44±0.88• | 23.99±0.47• | 13.28±0.73• | 7.37±0.39 |
| *genbase* | 0.36±0.09 | 0.53±0.15 | 0.86±0.41• | 0.39±0.09 | 0.65±0.23 | 0.27±0.02 |
| *medical* | 4.29±0.51• | 3.93±0.38• | 1.87±0.21 | 4.43±0.71• | 2.70±0.30• | 1.16±0.15 |
| *arts* | 9.47±0.35• | 6.08±0.26• | 4.95±0.22• | 8.68±0.31• | 5.68±0.12• | 3.69±0.18 |
| *business* | 6.13±0.52• | 2.59±0.27• | 2.15±0.15• | 5.71±0.39• | 2.25±0.15• | 1.32±0.13 |
| *education* | 13.76±0.89• | 4.68±0.22• | 3.84±0.28• | 9.84±0.62• | 3.94±0.32• | 2.52±0.21 |
| *entertain.* | 7.81±0.30• | 3.58±0.16• | 3.09±0.03• | 6.29±0.69• | 3.39±0.15• | 2.13±0.10 |
| *health* | 9.49±0.68• | 3.54±0.18• | 2.77±0.12• | 6.89±0.98• | 3.28±0.17• | 1.69±0.14 |
| *science* | 15.23±0.43• | 8.66±0.25• | 6.14±0.15• | 13.22±0.79• | 7.14±0.15• | 3.70±0.16 |
| *social* | 9.75±0.61• | 4.54±0.37• | 3.10±0.18• | 6.91±0.54• | 3.60±0.39• | 1.47±0.19 |

Table 3(e) Performance of different algorithms in terms of *Average precision*. (mean±std. %)

| Dataset | RAKEL | IBLR-ML | CLR | CChains | MLKNN | VPCME |
|---|---|---|---|---|---|---|
| *yeast* | 70.82±1.54• | 76.72±1.50• | 74.70±1.58• | 63.06±0.90• | 76.47±1.55• | 80.41±0.73 |
| *scene* | 82.67±0.82• | 86.68±0.70• | 82.64±1.01• | 71.33±1.91• | 86.25±0.63• | 90.32±0.89 |
| *enron* | 60.89±2.06• | 60.95±1.94• | 70.25±1.65• | 57.02±0.40• | 62.51±1.22• | 72.86±0.66 |
| *genbase* | 99.09±0.55 | 98.32±0.69• | 98.60±0.71 | 98.99±0.47 | 98.70±0.45 | 99.32±0.35 |
| *medical* | 83.07±2.63• | 74.18±0.95• | 87.58±1.32• | 83.61±2.67• | 80.57±2.50• | 89.83±1.06 |
| *arts* | 47.72±0.54• | 49.15±1.94• | 56.86±1.43• | 46.55±1.68• | 50.05±1.80• | 63.41±0.41 |
| *business* | 81.44±1.12• | 87.25±1.23• | 87.56±1.26• | 78.04±0.77• | 87.96±0.94• | 90.65±0.59 |
| *education* | 45.76±1.78• | 53.29±0.83• | 56.67±0.72• | 45.43±0.99• | 55.74±0.97• | 65.93±0.82 |
| *entertain.* | 54.39±1.16• | 58.83±0.98• | 63.05±0.58• | 53.68±2.55• | 59.53±1.37• | 71.54±1.14 |
| *health* | 67.42±1.60• | 68.96±1.37• | 74.73±1.26• | 65.45±2.33• | 68.56±1.45• | 78.58±0.87 |
| *science* | 41.97±0.72• | 43.33±1.14• | 52.41±1.33• | 40.74±1.04• | 45.91±1.26• | 58.83±1.16 |
| *social* | 66.34±1.58• | 66.20±1.01• | 73.25±1.21• | 67.16±1.38• | 68.89±1.42• | 77.40±1.08 |

(2) CChains performs worse than others, which might be due to fact that the order of the classifier chains in iteration is inappropriate for the datasets. RAKEL strives to learn a label power set classifier for each *k*-subset of labels, but the divided subsets are randomly selected from the dataset, which might degrade its performance.

(3) IBLR-ML outperforms RAKEL and CChains, since it combines logistic regression into one unified framework. However, the biased estimations of optimal regression coefficients usually lead to the imbalance between the global and local inference, which has negative effects on the performance.

(4) CLR performs better on text datasets than other algorithms except VPCME. Recall that CLR is a label ranking method and its artificial calibration label determines the separating boundary between relevant and irrelevant labels. As a result, if the confidence of the key calibration label is far from the desired, CLR tends to be outperformed by others.

Table 4 The win/tie/loss results for VPCME against the compared algorithms

| Evaluation Metrics | The VPCME method against | | | | |
|---|---|---|---|---|---|
| | RAKEL | IBLR-ML | CLR | CChains | MLKNN |
| *Hamming loss* | 7/4/1 | 8/4/0 | 7/4/1 | 10/1/1 | 5/7/0 |
| *Ranking loss* | 11/1/0 | 12/0/0 | 12/0/0 | 11/1/0 | 12/0/0 |
| *One-error* | 11/1/0 | 10/2/0 | 10/2/0 | 11/1/0 | 10/2/0 |
| *Coverage* | 11/1/0 | 9/3/0 | 11/1/0 | 11/1/0 | 9/3/0 |
| *Ave. precision* | 11/1/0 | 12/0/0 | 11/1/0 | 11/1/0 | 11/1/0 |
| In total | 51/8/1 | 51/9/0 | 51/8/1 | 54/5/1 | 47/13/0 |

(5) MLKNN is compared as a baseline here, since it is selected as the base classifier of our VPCME framework. Typically, it is a binary relevance learner, which implements each individual classifier through a combination of KNN and Bayesian inference [47]. Moreover, it often exhibits more simplicity compared to RankSVM and AdaBoost.MH [9].

(6) The disparity degrees of the six algorithms are somewhat tiny in terms of *Hamming loss* on the text datasets. This evidence demonstrates that the smaller *cardinality* and the lower *density* lead to the robustness of this metric, since the hit ratio of predictive label and ground-truth label becomes very large to some extent.

### 4.4.2 Results of Group2

As illustrated in Fig.1, the performance of our proposed method with different variable pairwise constraint thresholds is vividly depicted.

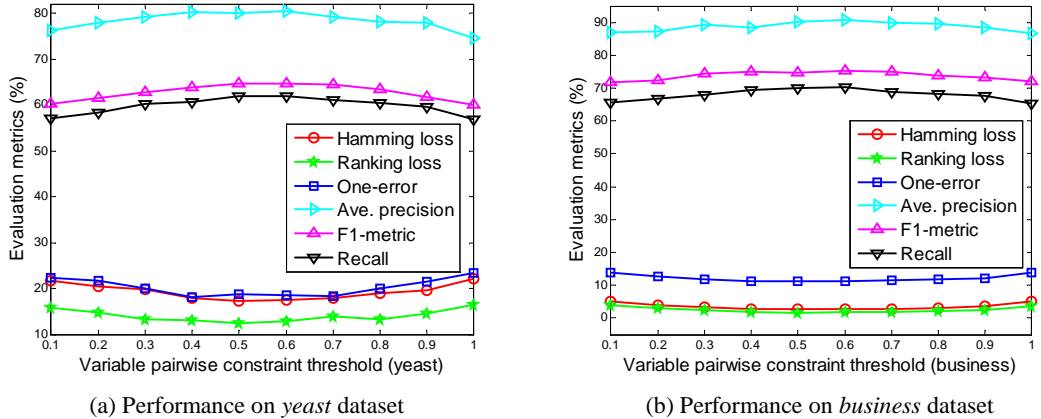

(a) Performance on *yeast* dataset  (b) Performance on *business* dataset

Fig.1. Performance of VPCME under different thresholds

On the left panel, Fig.1 (a) reflects the comparison performance on the biology data *yeast*. On the right panel, Fig.1 (b) displays the behavior of VPCME on the text data *business*. From the figures, it can be seen that in terms of *Average precision*, *F1-metric* and *Recall*, the curves of VPCME begin to ascend at the initial stage and keeps relatively stable during the intermediate period (i.e., 0.4~0.7), but it declines as the threshold approaches one. Obviously, VPCME behaves differently compared to *Hamming loss*, *Ranking loss* and *One-error*.

Clearly from the two figures, we find that the tendency on the left is more significant than that on the right, which indicates our method enjoys robustness over a larger range on the text data. We attribute this to the fact that the imbalance problem caused by the ratio of *cannot-link* constraints and

*must-link* constraints should be neither too large nor too small, since extreme values will seriously break the balance between them. Herein, the curves reflect that our framework tends to perform well in a wide range of varied thresholds.

Table 5(a) Performance of VPCME with different ensemble sizes on *medical.* (mean±std. %)

| Evaluation Metrics | $S = 1$ | $S = 10$ | $S = 20$ | $S = 30$ | $S = 40$ | $S = 50$ |
|---|---|---|---|---|---|---|
| *Hamming loss*↓% | 2.74±0.16 | 2.02±0.18 | 1.46±0.14 | 1.25±0.08 | 1.24±0.11 | 1.17±0.09 |
| *Ranking loss*↓% | 2.82±0.25 | 2.09±0.35 | 1.93±0.28 | 1.70±0.21 | 1.61±0.19 | 1.56±0.26 |
| *One-error*↓% | 18.39±1.44 | 17.13±1.27 | 16.54±1.39 | 16.16±1.56 | 16.08±1.33 | 15.98±1.21 |
| *Coverage*↓ | 2.15±0.17 | 1.95±0.24 | 1.47±0.19 | 1.16±0.15 | 1.11±0.26 | 1.04±0.12 |
| *Ave. precision*↑% | 86.21±1.09 | 88.54±1.23 | 89.19±1.15 | 89.93±1.06 | 90.04±0.95 | 90.15±1.16 |

Table 5(b) Performance of VPCME with different ensemble sizes on *entertainment.* (mean±std. %)

| Evaluation Metrics | $S = 1$ | $S = 10$ | $S = 20$ | $S = 30$ | $S = 40$ | $S = 50$ |
|---|---|---|---|---|---|---|
| *Hamming loss*↓% | 6.83±0.17 | 6.41±0.21 | 5.93±0.16 | 5.62±0.14 | 5.54±0.10 | 5.36±0.12 |
| *Ranking loss*↓% | 9.05±0.38 | 7.99±0.44 | 7.47±0.31 | 7.02±0.39 | 6.87±0.28 | 6.74±0.34 |
| *One-error*↓% | 42.86±0.92 | 40.25±1.13 | 39.79±1.02 | 39.36±0.82 | 39.14±0.94 | 39.08±0.85 |
| *Coverage*↓ | 3.37±0.19 | 3.06±0.11 | 2.65±0.18 | 2.13±0.10 | 2.09±0.08 | 1.96±0.14 |
| *Ave. precision*↑% | 67.24±1.45 | 69.92±1.37 | 70.86±1.05 | 71.54±1.14 | 71.79±1.21 | 71.95±1.06 |

### 4.4.3 Results of Group3

To further examine the proposed method, we show the performances of VPCME under different ensemble sizes in Table 5. In these tables, Table 5 (a) summarizes the results on the *medical* data and Table 5 (b) makes records of the *entertainment* data.

The results clearly show that our proposed method consistently outperforms others as ensemble size increases. In particular, when the single MLKNN is used (i.e., S=1), VPCME still has good performance in the new data space. However, when S exceeds 30, the performance is improved very slightly. This implies that the ensemble performance will achieve a stable value after some peak value. Generally, we should not choose a larger ensemble size than some proper threshold, since training more base classifiers would cost much more time and memory.

### 4.4.4 Results of Group4

As vividly depicted in Fig. 2, it is readily to see that the proposed VPCME framework consistently performs better than other compared approaches in terms of various evaluation criteria. Particularly, we find that AdaBoost$_{PDT}$ performs worst since it does not consider the correlations among multiple labels. Bagging$_{VPCP}$ performs better than AdaBoost$_{PDT}$ and MDDM, because it exploits the variable pairwise constraints of samples. But it is outperformed by VPCME, which is due to the fact that the Bagging$_{VPCP}$ framework treats every sample equally while the VPCME framework puts more emphasis on the misclassified samples in iteration. In addition, MDDM is a dimensionality reduction method via maximizing the dependence between the original features and the associated class labels, and it does not take into account the pairwise constraints.

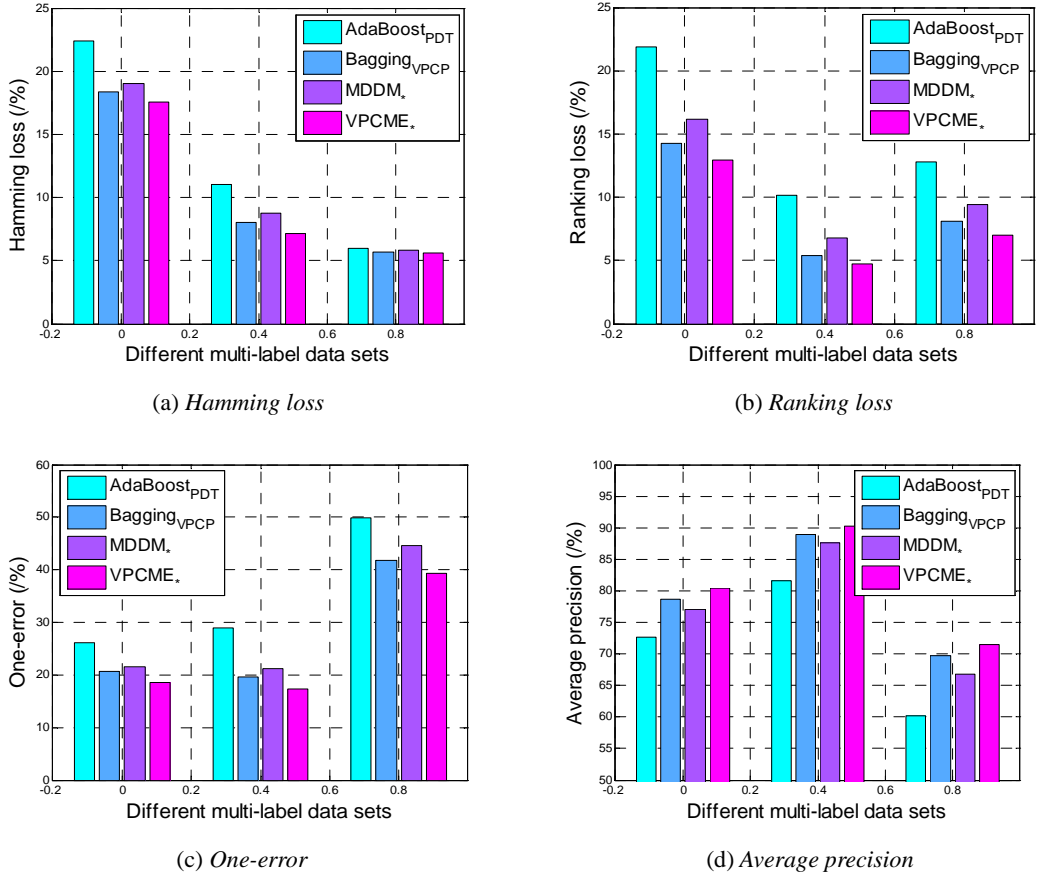

Fig.2 Comparison of different approaches in terms of four metrics

## 5 Conclusions

In this paper, we introduce a novel multi-label classification framework called *Variable Pairwise Constraint projection for Multi-label Ensemble* (VPCME) to construct a multi-label ensemble for handling multi-label data. This framework involves two inherent components, i.e., the variable pairwise constraint projection and the boosting-like strategy. In detail, we employ the variable pairwise constraint projection to obtain a well preserved lower-dimensional data space, where the base classifiers are learned. Besides, we make use of a boosting-like strategy to improve the generalization ability of the classifier. For the boosting-like strategy, both the variable pairwise constraints and the bootstrap steps are exploited to diversify a group of base classifiers. In this work, the *majority voting* is adopted to decide the estimated label set for each test data. We conducted extensive interesting experiments over a range of multi-label datasets. Results show that our proposed approach performs better than other competing methods.

Nevertheless, there still remain some problems to be explored in future. For example, it is sensible to develop one principle way to select the optimal threshold for variable pairwise constraints. Due to the high dimensions of many real-world data, it is practically important to study the joint learning of multi-label feature selection and multi-label ensemble so as to select the most informative feature subsets. Another interesting problem is to explore the way to speed up the multi-label ensemble approach such that the computational costs can be greatly reduced.

## Acknowledgements

We would like to thank the anonymous reviewers and Dr. Zhang Lijun for their insightful comments and suggestions, which greatly help improve this paper. This work was supported in part by the National Science Foundation for Outstanding Young Scientists of China (60425310) and the Fundamental Research Funds for the Central Universities (2012FZA5017).